# Using massive health insurance claims data to predict very high-cost claimants: a machine learning approach

**Authors:**

José M. Maisog[1,¶], Wenhong Li[1,&], Yanchun Xu[1,&], Brian Hurley[1,&], Hetal Shah[1,&], Ryan Lemberg[1,&], Tina Borden[1,&], Stephen Bandeian[1,&], Melissa Schline[1,&], Roxanna Cross[1,&], Alan Spiro[1,&], Russ Michael[1,¶], Alexander Gutfraind[1,2,¶,*]

[1] Blue Health Intelligence, Chicago, Illinois, United States of America.

[2] University of Illinois School of Public Health & Loyola University of Chicago

[*] Corresponding author.

E-mail: sasha.gutfraind@bluehealthintelligence.com and agutfraind.research@gmail.com (AG)

[¶] These authors contributed equally to this work.

[&] These authors also contributed equally to this work.

# Abstract

Due to escalating healthcare costs, accurately predicting which patients will incur high costs is an important task for payers and providers of healthcare. High-cost claimants (HiCCs) are patients who have annual costs above $250,000 and who represent just 0.16% of the insured population but currently account for 9% of all healthcare costs. In this study, we aimed to develop a high-performance algorithm to predict HiCCs to inform a novel care management system.

Using health insurance claims from 48 million people and augmented with census data, we applied machine learning to train binary classification models to calculate the personal risk of HiCC. To train the models, we developed a platform starting with 6,006 variables across all clinical and demographic dimensions and constructed over one hundred candidate models. No requirement regarding continuous insurance coverage was imposed on the training and holdout datasets.

The best model achieved an area under the receiver operating characteristic curve of 91.2%. The model exceeds the highest published performance (84%) and remains high for patients with no prior history of high-cost status (89%), who have less than a full year of enrollment (87%), or lack pharmacy claims data (88%). It attains an area under the precision-recall curve of 23.1%, and precision of 74% at a threshold of 0.99. A care management program enrolling 500 people with the highest HiCC risk is expected to treat 199 true HiCCs and generate a net savings of $7.3 million per year.

Our results demonstrate that high-performing predictive models can be constructed using claims data and publicly available data alone, even for rare high-cost claimants exceeding $250,000. Our model demonstrates the transformational power of machine learning and artificial intelligence in care management, which would allow healthcare payers and providers to introduce the next generation of care management programs.

# Introduction

It has been known for decades that a relatively small group of patients, termed high-cost claimants (HiCCs), accounts for a disproportionate share of healthcare costs and insurance claims [1]. For example, members with claims over $50,000 per year represented 1.2% of the U.S. insured population but comprised 31% of total spending [2]. In the Medicare population in the U.S., McWilliams and Schwartz [3] found that 17% of the population incurred 75% of all costs. In our comprehensive data on the non-Medicare insured population [4], members with annual costs greater than $250,000 comprise just 0.1% of the population yet account for 9% of overall costs. Moreover, in this population the number of HiCCs with $250,000 or more has risen by 62% from 36,449 in 2012 to 58,897 in 2016, and the average cost was $446,748 per HiCC in 2017. It is therefore not surprising that when asked to list the most important strategies for healthcare in the next five years, midsized and large employers ranked managing high-cost claimants at the top of the list [5].

As a population, HiCCs are frequently burdened with multiple chronic diseases, functional limitations, and other barriers [6]. Often, high medical expenditures occur as part of acute or invasive therapies that are frequently unsuccessful and involve tremendous suffering and disability [7]. Fortunately, there are many interventions that could prevent relatively healthy individuals from becoming high-cost claimants in the first place [8,9]. An intervention is tailored to a member's specific circumstances and might consist of some combination of a telephone call, additional diagnostic studies, referral to a specialist, digital coaching, or other services. Health insurers and providers often maintain a care management organization whose role is to identify patients who would benefit from interventions and match them with intervention programs (Fig. 1).

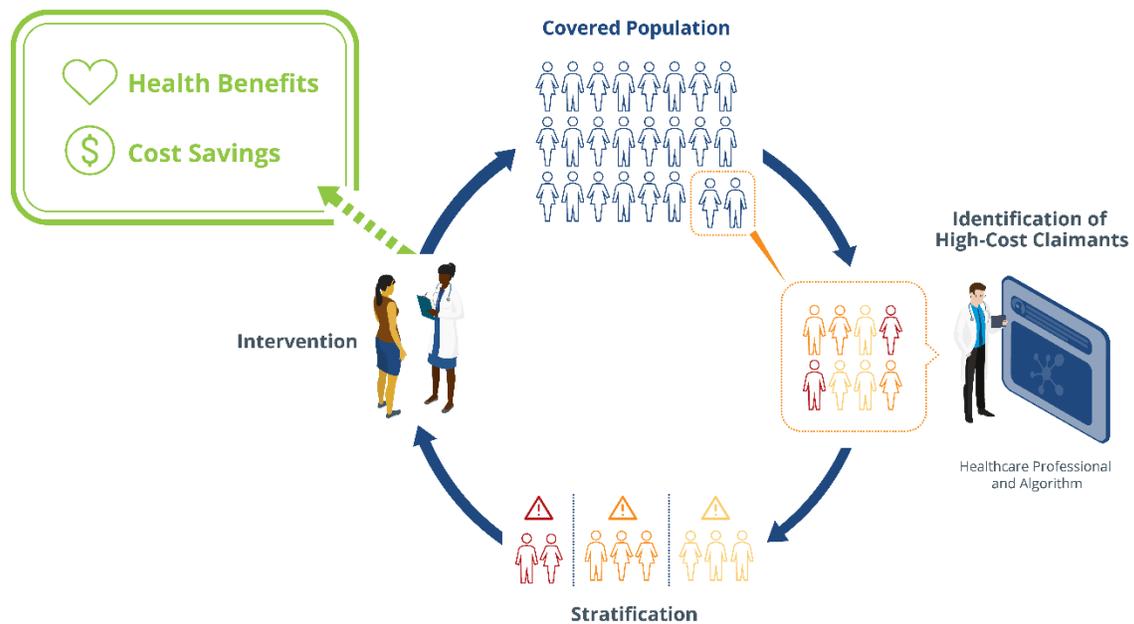

**Figure 1. The care management process**. Members in the covered population are first reviewed by the healthcare professional, who, informed by the algorithm, selects some of the members for possible intervention. Selected members are then stratified according to dimensions such as risk and the availability of suitable interventions, and if appropriate, receive an intervention. This referral decision is not based solely on the algorithm's prediction score; rather, the decision is holistic, with the algorithm's prediction score to be considered in the context of the member's overall clinical situation. The intervention is expected to result in some health benefits and cost savings. After the intervention the individual returns to the population.

Because interventions are costly, potential benefits and cost savings that might result from the intervention must be balanced against the cost of the intervention itself [10]. Furthermore, it is challenging to determine which individuals would derive the greatest benefit from the intervention, even within a narrow subpopulation that has a serious pre-existing condition. Therefore, one of the central challenges of a care management program is to identify individuals at risk of acute and expensive health outcomes [8].

The emergence of powerful predictive AI methods seems ideally suited to address this identification challenge, since an AI algorithm could potentially predict future costs or medical needs at an individual level [11,12]. Such an identification algorithm can then guide limited intervention resources towards the highest-risk and highest-need individuals. Therefore, our goal here was to apply machine learning to identify members who are at risk of exceeding a total healthcare cost of $250,000 over the next 12 months. We hypothesized that using a relatively large dataset of insurance claims and a large investment in engineering new input variables we could exceed previously published benchmarks in this field.

## Overview of existing research

Much research has examined the problems of predicting medical costs and identifying high-cost claimants [12–14]. However, not many studies used very high-cost thresholds, i.e., studies examining the top 1% (or higher) of claimants. Additionally, many more studies did not report predictive performance in terms of area under the ROC or precision-recall curves, or were descriptive in nature rather than predictive; see **Table 1**.

Historically, the problem of high-cost claimants was studied by actuarial scientists at the population level, with emphasis on parameter estimation and statistical significance tests [15–19]. But with the growing availability of data, computing power, and new artificial intelligence and machine learning methods (AI/ML), there has recently been increased interest in predicting costs at the level of the individual member instead of estimating parameters for populations [13,20] (for general treatments on the contrast between model- and data-driven methods see [21–23]).

**Table 1**: Summary of studies that used machine learning methods to identify high-cost claimants. Studies were selected by filtering the list of 55 studies in Table 1 of [14], and retaining only those studies where the population to be identified was the top 1% or rarer. High-cost claimants above $250,000 have a nearly 100-fold lower prevalence than the top 10% of claimants, posing a much greater predictive challenge. The only study that reported performance values was [24] (area under the ROC curve, AUC-ROC: 81%-86%) and all other studies at this high threshold either did not assess predictive modeling and/or did not report the AUC-ROC.

| Study | Population to be identified | Model/algorithm |
| --- | --- | --- |
| Ash et al. 2001 [25] | Top 0.5% | Logistic regression |
| Coughlin & Long 2010 [26] | Top 1%, top 5%, top 10%, top 25%, top 50%, and bottom 50% | Descriptive only |
| DeLia 2017 [27] | Top 1%, top 10%, and bottom 90% | Multinomial probit model |
| Hensel et al., 2016 [28] | Top 1%, top 2%-5%, top 6%-50%, bottom 50%, and a zero-cost nonuser group | Logistic regression |
| Meenan et al., 2003 [29] | Top 0.5% | Risk modeling (linear regression) |
| Monheit, 2003 [30] | Top 1%, top 5%, top 10%, top 20%, top 30%, top 50%, and bottom 50% | Logistic regression |
| Powers and Chaguturu, 2016 [31] | Top 1% | Descriptive only |
| Riley, 2007 [32] | Top 1% and top 5% | Descriptive only |
| Robst, 2015 [33] | Top 1% | Logistic regression |
| Rosella et al., 2014 [34] | Top 1%, top 2-5%, top 6%-50%, and bottom 50% | Multinomial logistic regression |
| Wammes et al., 2017 [35] | Top 1% and top 2%-5% | Descriptive only |

| Wodchis et al., 2016 [36] | Top 1%, top 5%, top 10%, and top 50% | Descriptive only |
| Zhao et al., 2003 [37] | Top 0.5% | Linear regression |

## Detailed review of previous prediction approaches

A common approach to the problem of predicting high-cost claimants is the use of logistic regression. A 2010 paper using logistic regression found that inclusion of medical condition information substantially improved the prediction of high-cost patients, resulting in "good discrimination" (area under the ROC curve, AUC-ROC=0.84%) [38]. The paper also concluded that the number of chronic conditions should be considered as a predictor for high-cost prediction models. In a 2015 study, logistic regression was used to predict which patients would transition from an intermediate-cost subpopulation to a high-cost subpopulation with "reasonable discrimination" (AUC-ROC=0.67%) [39]. Two predictors that were significantly associated with high future costs were the count of chronic conditions and having a diagnosis of congestive heart failure. In a 2017 paper, a group-based trajectory model based on logistic regression was applied to data from a large insurer to accurately predict patients in the highest spending trajectory and the top fifth percentile for spending [40]. Using data from the Danish National Health Service and Civil Registration System, Tamang et al. implemented a penalized logistic regression model with over 1,000 predictors and were able to achieve good predictive performance (AUC-ROC of 0.79%) in a "cost bloom analysis" [41].

Alternative methods such as machine learning techniques are being applied increasingly to the problem of predicting high-cost claimants. Using neural networks, a 2005 study comparing a population model against three disease-specific models found that larger cohorts tended to result in a greater predictive power of the disease-specific models compared to the population model [42]. A 2013 study using routine electronic service records found that a score based on six simple dichotomous questions had only "fair" predictive power for health and social care costs in elder patients discharged from acute medical units, with an AUC-ROC of 0.70% [43]. A Canadian group [44] applied machine learning to a large range of clinical measurements to identify the top 5% of claimants, attaining an AUC-ROC of 0.81%-0.94%. In a study where an extended gradient boosting model (XGBoost [45]) was applied on an imbalanced dataset from three of the largest health insurers in the U.S., Hartman found that oversampling the minority class resulted in better predictive performance (AUC-ROC 0.835) than undersampling the majority class [46], at least at the highest thresholds. Gibbs et al. proposed the use of asymmetric cost matrices to optimize the threshold for an intervention [10].

Whether one uses logistic regression or alternative machine learning techniques, obvious questions might be (1) which analytic approaches tend to result in the best predictive performance? and (2) which predictors tend to best explain high utilization? A recent review of 55 papers in the literature revealed that high utilization was primarily explained by high levels of chronic and mental illness [14]. Another 2018 literature review concluded that gradient boosting had the best predictive performance overall and for low- to medium-cost members, but neural networks and ridge regression had the highest performance for high-cost members [13].

## Our contribution

In this paper, we describe a novel solution to the problem of identifying future high-cost claimants using machine learning (ML). We applied one of the largest datasets of healthcare insurance claims (over 50 million members), achieving one of the highest performance results reported in the literature reviewed here. Because health insurance plans typically have access only to their own claims data and not to hospital records or specialized registries, we constrained our algorithm to use variables available through health insurance claims alone, along with public data on social determinants of health. No hospital records or specialized registries were used.

Because we anticipated using the predictive model on the full population without any filtering, no requirement regarding continuous enrollment in health insurance was imposed on the data. The algorithm was therefore designed to operate for members with an incomplete medical history, such as less than one year of data. Additionally, the algorithm can function for members for which only medical claims and not pharmaceutical claims are available – a situation occurring in approximately 40% of the insured population in our data.

Similar previous studies most often defined HiCCs using thresholds between $50,000 and $100,000 per year; some other studies used instead the top 1% to 10% in costs [14]. We instead defined HiCCs as members with a yearly total allowed amount that exceeded $250,000: this is the threshold at which many reinsurance policies attach. At this threshold, high-cost claimants accounted for about 1.6 out of 1,000 members, making identification particularly challenging and requiring truly massive datasets.

Our goal was to identify members who would exceed a total healthcare cost of $250,000 over the next 12 months. Generally, identification of high-cost claimants is modeled typically as a prediction problem in the framework of machine learning. Two formulations are used: (1) predicting cost, namely predicting a member's future dollar cost amount over the next 12 months; or (2) binary classification, namely, predicting whether or not a member will exceed a certain cost amount. We evaluated both of these formulations and then selected the second formulation (classification) after considering the business requirements that are often around a particular threshold and the performance of the models.

Applying our methods to the problem we found that the best performing model overall was the Light Gradient Boosted Trees Classifier (LightGBM) algorithm [47], achieving an AUC-ROC of 91.25% on a holdout dataset. This is consistent with the findings of a recent literature review [13], where gradient boosting methods had the best predictive performance overall. The model's AUC-ROC was 7% higher than all previous models at the $250,000 point, and it is estimated that the model's high performance could generate considerable benefits for care management programs.

## Methods

### Construction of input variables

Our model for predicting health costs was constructed from administrative claims data – the data created as part of electronic exchanges between medical facilities, professionals, and pharmacies on the one side, and payers on the other. In the U.S. the format of claims data is standardized by HIPAA and contains a listing of diagnoses, procedures, drugs, as well as costs. Taken together, claims data provide a nearly complete summary of each patient's medical journey across all types of care. Claims data are readily available to all actors in the healthcare system – the medical insurers, government payers, and sponsors of healthcare – and are increasingly available to patients.

The majority of input variables were calculated from claims data over the one-year report period 4/1/2017 to 3/31/2018, called the "reporting period." The remaining variables, specifically cost trend and some enrollment variables, also included any claims available before the start of the reporting period. If no data were available, we assigned null values. The following one-year interval from 4/1/2018 to 3/31/2019 was considered the "prediction period" and was used to define the high-cost status. At the time of the calculation, claims data were >99% adjudicated.

Predictors were constructed using SQL (Vertica Analytic Database v8.0.1-3; Micro Focus, Berkshire, England, UK). Total allowed amounts for medical conditions (diagnoses) and procedures (services) in the current report period were summed using BHI's Episodes of Care grouper methodology [48]. To prevent information leakage, data were processed in an isolated schema that excluded any events occurring outside the report period. Indicators of social determinants of health were constructed using publicly available data from the U.S. census, estimated for 2017 American Community Survey (ACS) 5-year estimates at the ZIP code level, which were linked to the member based on the member's ZIP code. ACS data included housing conditions, unemployment, poverty and fraction of the population who are a racial minority.

Table 2 lists the types of variables used in the prediction. The variables were selected by pruning an initial list of 6,006 variables, which improved the performance. Feature pruning was performed using a feature importance metric available in DataRobot, which implemented a model-agnostic algorithm based on permutation testing. We also used a variable importance metric available in DataRobot, which captured information about tree splitting in tree-based algorithms.

**TABLE 2**. List of the types of input variables. The list of input variables used in the final model were selected from a larger list of 6,006 variables. The total number of predictors before and after feature selection was 6,006 and 255, respectively. Columns "Original Count" and "Final Count" show the number of predictors before and after feature selection, respectively. Details are found in S1 Table.

| Variable Type | Group | Original Count | Final Count | Source | Explanation |
|---|---|---|---|---|---|
| Personal Risk Factors | Personal data | 32 | 15 | Claims | Age, gender, family size, industry, and others |
| | Eligibility and enrollment coverage | 17 | 5 | Claims | Insurance coverage, e.g., number of days a member had coverage of a certain type |
| | Social determinants of health | 55 | 39 | Public data | Social vulnerability, education, poverty, minority status, English-speaking ability, housing, transportation, and others |
| Clinical history | Procedures, diagnoses, and drugs | 5,799 | 100 | Claims | The 12-month cost associated with a given procedure, diagnosis, or class of drugs |
| | Clinical events | 63 | 63 | Claims | Groups of ICD-10 and CPT codes associated with high cost; count of events by type (Emergency Department, ambulatory, and hospital inpatient) |

|  | Derived indices | 3 | 3 | Claims | Mortality rate, actuarial life expectancy, and years of life lost |
| --- | --- | --- | --- | --- | --- |
| Cost history | Cost indices | 26 | 19 | Claims | Cost over the past 12 and 24 months, total cost in current and past 12 months, cost of specialty drugs, days of drug supply, and others |
|  | Cost trends | 11 | 11 | Claims | Changes in cost over 12-, six- and three-month intervals, predicted 12-month cost, cost trends, and others |

## Training and holdout population

The model was trained and tested using realistic data, since we anticipated using the predictive model on the entire insured population including individuals with incomplete medical histories. Members were required to be enrolled only on the first day of the last month of the reporting period (4/1/2017 to 3/31/2018) and on the last day of the first month of the predicted period (4/1/2018 to 3/31/2019). No requirement regarding continuous enrollment was imposed on the data. In the resulting population, 63% had both medical and pharmaceutical claims, and 78% had at least one year of continuous enrollment.

Due to limits on the file size for upload to DataRobot, the training dataset was reduced to a subpopulation of 3 million members by downsampling the majority class as follows [46]. Due to the high threshold of $250,000, HiCCs are rare and therefore each becames very informative. All high-cost claimants in the original training dataset were retained. The non-HiCCs in the original training dataset were randomly sampled to bring the total complement in the reduced training dataset to 3 million. In the final training dataset, the number of HiCCs and non-HiCCs was 61,277 and 2,938,723, respectively, and 20% of the training data was used for model selection. Training of all candidate models was completed in 5 hours using a high-performance cluster.

To report on model performance, the final model was evaluated on a holdout dataset of 9,684,279 members, described in Tables 3 and 4, which was not used in the training. The proportion of high-cost claimants (HiCCs) in the holdout reflected the natural proportions in the commercially insured population in the US. A small fraction of the population 65 or older is also was included in this population. We reported the performance on all age and gender groups. Additionally, we further stratified the HiCCs into emergent and recurrent, corresponding to individuals with no previous history of HiCC status, and those who previously were high-cost claimants. These were, in effect, two distinct prediction problems: the emergent population was identified from a very large (N= 48,402,958) set of candidates, whereas the recurrent population was identified in a much smaller set (N= 28,249), of which approximately 37% became HiCCs in subsequent years.

**TABLE 3.** Shown are demographics of the holdout population used for model evaluation.

|  | Population | Members | HiCCs | HiCCs |
|---|---|---|---|---|
|  | Total holdout | 9,684,279 | 15,051 | 0.16% |
| Gender | Female | 4,884,419 | 7,068 | 0.14% |
|  | Male | 4,799,860 | 7,983 | 0.17% |
| Age group | 0-17 | 2,066,880 | 1,085 | 0.05% |
|  | 18-64 | 7,389,332 | 12,490 | 0.17% |
|  | 65+ | 228,067 | 1,476 | 0.65% |
| HiCC history | Emergent | 9,678,645 | 12,973 | 0.13% |
|  | Recurrent | 56345,634 | 2,078 | 36.9% |

**TABLE 4.** Shown are the highest-cost conditions for HiCCs (same year, top 20 categories ordered by number of HiCCs), in the holdout population. * indicates not otherwise specified.

| Highest-cost condition | Number of members | Proportion of all HiCCs who have the condition | Proportion of all members with the condition who are HiCC |
|---|---|---|---|
| Cancer* | 3,762 | 1.3% | 5.9% |
| Coronary artery disease | 2,109 | 0.8% | 1.1% |
| Pneumonia | 1,552 | 0.6% | 1.6% |
| Renal failure - acute | 1408 | 0.5% | 6.7% |
| Heart failure | 1,397 | 0.5% | 4.5% |
| Renal failure - chronic | 1,313 | 0.5% | 11.5% |
| Cancer - breast | 1,033 | 0.4% | 2.6% |
| Sepsis | 751 | 0.3% | 4.6% |
| Respiratory failure | 705 | 0.3% | 5.9% |
| Cancer - lung | 704 | 0.3% | 13.9% |
| Polyneuropathy | 534 | 0.2% | 2.4% |
| Endocrine disorder* | 521 | 0.2% | 1.9% |
| Cancer – multiple myeloma | 483 | 0.2% | 25.2% |
| Cancer – colon* | 464 | 0.2% | 5.9% |
| Regional enteritis | 365 | 0.1% | 1.2% |
| Congenital coagulation disease | 285 | 0.1% | 10.2% |
| Cancer - brain | 208 | 0.1% | 11.2% |
| Cancer - acquired hemolytic anemia | 133 | 0.05% | 13.1% |
| Cancer - acute lymphoid leukemia | 111 | 0.04% | 10.6% |
| Cancer - acute myeloid leukemia | 103 | 0.04% | 13.8% |

## Machine learning and statistical methods

BHI uses DataRobot's predictive platform (DR) version 5.0.1 (DataRobot, Boston, MA). Our platform includes industry-leading tools for exploratory data analysis (EDA), model training, validation, and deployment. Once the training data were uploaded, dozens of machine learning models were trained in a high-throughput supervised learning system. Training used the log-loss error function and all the classifiers have built-in regularization to minimize overfitting in the presence of class imbalances (see e.g. [49]). Missing values are extremely rare and were imputed automatically with median values. Data used in model training were stripped of basic HIPAA identifiers and anonymized. We have implemented multiple levels of security that governed access to both the models and the results.

Model selection proceeded in three rounds, where in each round, more data were used (16%, 32%, and then 64% of the sample), and the best-performing algorithm in each round was passed on to subsequent rounds. The final round included a grid scan of hyperparameters. The algorithms considered included Random Forests, Support Vector Machines, Gradient Boosted Trees, Elastic Nets, Extreme Gradient Boosting, and ensembles. Multiple implementations of the algorithms were tested, including the open source machine learning libraries from R, scikit-learn, TensorFlow, Vowpal Wabbit, Spark ML, XGBoost, and LightGBM.

Once the automatic selection was complete, we reviewed the model performance on a holdout dataset. The top model was selected by reviewing each model holistically, including its predictive performance, scoring speed, and interpretability. The model was then subjected to a clinical review and assessment in a separate holdout dataset (see below). We found that we could improve the model's performance by pruning variables of lower importance. We minimized the risk of overfitting by preferring algorithms that are inherently resistant to overfitting such as LightGBM, and by confirming that the model's performance was consistent on the training and holdout data.

Because HiCCs are rare, we used performance metrics that are appropriate for classification problems with class imbalances [50,51]: area under the ROC Curve (AUC-ROC) and area under the precision-recall curve (AUC-PR). The AUC-PR has been increasingly suggested as the best overall metric of model performance [52], but we used both metrics since most of the existing literature in the field still uses the AUC-ROC metric. We also computed recall (also called sensitivity and true positive rate), precision (also called positive predictive value), false positive rate, F1 score, and Matthews correlation coefficient (MCC). In production, we applied isotonic regression to obtain calibrated probability scores [53]. The optimal prediction threshold was normally selected using economic analysis (see below) or in other cases, by selecting the score that maximizes the F1 score and the MCC – which agreed within 1% [54].

The model was deployed both as an API connected to our predictive platform and as an executable Java Archive file (JAR), which can be run in any environment supporting Java (e.g., Linux or Windows). The consistency of the scores of the two implementations was checked to agree numerically to within 0.0001. Model validation used JARs containing the model to calculate prediction scores. The running time of the JAR model is over 400,000 rows per minute. We applied a stringent development and validation process to ensure credibility and accuracy in our recommendations. Our development process and model training are very dependent on iterative clinician review and acceptance by clinical staff. Also, the internal model validation was conducted by an independent team.

## Health economics methods

Going beyond statistical methods for performance, we estimated the financial and health impact of the model. For this estimation, we placed the model in a typical scenario of a care management program in which the model identifies individuals in need of timely interventions [8]. In such a program, individuals with the highest model score are evaluated by clinical experts such as nurses, and if appropriate, enrolled in an intervention program. We compared the identification model to the common care management programs, that often use simple rules to identify members at risk, and therefore have a very high false alert rate.

Because the model is expected to be used in a care management program, the model's probable financial and health impact was assessed in the following care management scenario, which is closely based on our data from multiple health insurers in the U.S. In this scenario, one million members are covered by the insurer, the rate of HiCCs is 0.16%, and the mean cost per HiCC case is $413,975. We

assumed an intervention program costs $10,000 per member and achieves an average cost reduction of 15% cost reduction per HiCC. We conservatively did not attempt to estimate the value of the intervention beyond the first year or its effect on non-HiCC members. The care management program has the capacity to treat between 300 and 1000 people per year. The model was considered as a replacement for an existing rule-based HiCC identification system, which was assumed to have a precision of 2%.

## Results

Our model training found that highest-performing model was a Light Gradient Boosted Trees Classifier [47]. The classifier uses 410 trees with a maximum of 16 leaves per tree, boosted at a learning rate of 0.05, and with no regularization. The three most important predictor variables are age, a tendency for rising cost in the last three months of the prediction period, and life expectancy based on actuarial tables (see S1 Table).

When evaluated on the holdout dataset, the algorithm achieves an area under ROC curve of 91.2%, and an area under the precision-recall curve of 23.1% (Figure 2). At a threshold of 0.76 (consistent with the highest F1 score), the model gives recall of 33.0% and precision of 29.9% (see Table 5).

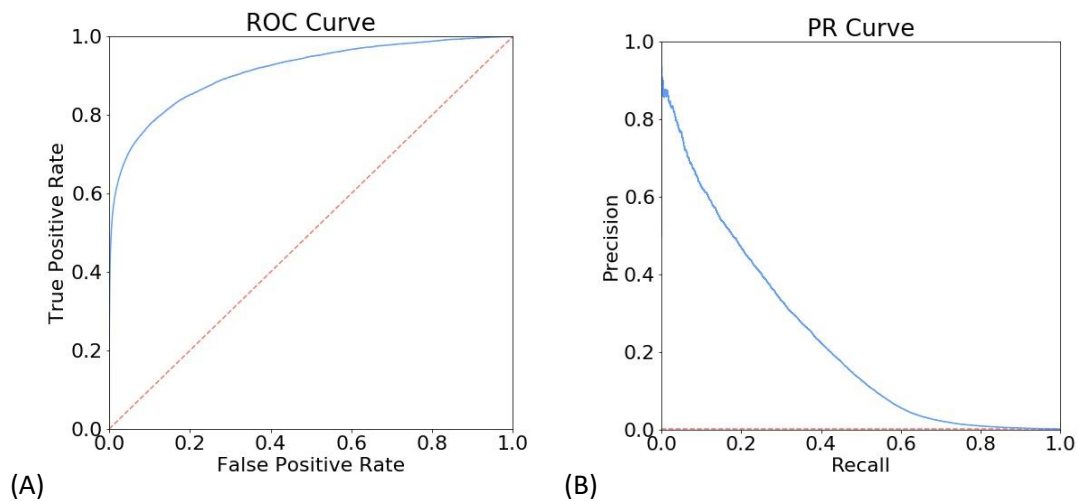

**Figure 2**: (A) The receiver operating characteristic of the HiCC predictive model is shown. It has an area under the ROC curve (AUC-ROC) of 91.2%. The red-dashed diagonal line indicates the chance-level performance benchmark. (B) The precision-recall curve of the HiCC predictive model is shown. It has an area under the PR curve (AUC-PR) of 23.1%. A red-dashed line just above the X axis indicates the reference, i.e., the proportion of high-cost claimants in the holdout data, or 0.16% (cf. TABLE 3). The model attains a precision > 80% when high predictive score thresholds are used.

**Table 5.** Shown is the performance of the model on the holdout population at various thresholds, including the threshold that maximized the F1-Score, 0.76. TP, FP, FN, and TN the number of true positives, false positives, false negatives, and true negatives, respectively. Recall is also called true positive rate or sensitivity; TNR is true negative rate (also called specificity); precision is also called positive predictive value; and NPV is negative predictive value.

| Threshold | TP | FP | FN | TN | Recall | TNR | Precision | NPV |
|---|---|---|---|---|---|---|---|---|
| 0.5 | 6553 | 28102 | 8498 | 9641126 | 43.54% | 99.71% | 18.91% | 99.91% |
| 0.6 | 6025 | 20930 | 9026 | 9648298 | 40.03% | 99.78% | 22.35% | 99.91% |
| 0.7 | 5436 | 15045 | 9615 | 9654183 | 36.12% | 99.84% | 26.54% | 99.90% |
| 0.76 | 4967 | 11632 | 10084 | 9657596 | 33.00% | 99.88% | 29.92% | 99.90% |
| 0.8 | 4606 | 9405 | 10445 | 9659823 | 30.60% | 99.90% | 32.87% | 99.89% |
| 0.9 | 3258 | 4049 | 11793 | 9665179 | 21.65% | 99.96% | 44.59% | 99.88% |
| 1 | 0 | 0 | 15051 | 9669228 | 0.00% | 100.00% | N.A. | 99.84% |

The following tables (Tables 6 and 7) show the model's performance in the population of emergent and recurrent HiCCs, namely those with and without prior history of HiCC status, respectively. The AUC of the model is higher in the emergent than the recurrent population (89% vs. 81%) and consistent between gender and demographic cohorts. Members that either (1) lacked data on pharmacy benefits or (2) lacked one full year of data represented 37% and 22% of the population, respectively, yet the model maintained its AUC in these populations to within 1%.

**TABLE 6.** Shown is the model performance in the emergent HiCC population. The threshold for a positive class is 0.76, which maximizes the F1 score. AUC is area under the ROC curve; recall is also called true positive rate or sensitivity; FPR is false positive rate; precision is also called positive predictive value; and NPV is negative predictive value.

| Population | AUC | N | Recall | FPR | Precision | NPV |
|---|---|---|---|---|---|---|
| National population | 88.72% | 48,402,958 | 24.93% | 0.11% | 16.63% | 99.93% |
| National prior year male population | 88.75% | 23,989,683 | 24.68% | 0.12% | 16.79% | 99.93% |
| National prior year female population | 88.62% | 24,413,275 | 25.21% | 0.11% | 16.45% | 99.94% |
| National prior year age group 0-17 population | 85.28% | 10,333,528 | 22.33% | 0.02% | 27.09% | 99.97% |
| National prior year age | 87.89% | 36,928,635 | 24.78% | 0.12% | 16.68% | 99.93% |
| group 18-64 population | | | | | | |
| National prior year age group 65+ population | 85.64% | 1,140,795 | 29.16% | 0.61% | 12.44% | 99.79% |
| National prior year pharmacy benefit population | 89.26% | 30,508,394 | 27.41% | 0.13% | 16.82% | 99.93% |
| National prior year no pharmacy benefit population | 87.67% | 17,894,564 | 20.05% | 0.08% | 16.13% | 99.94% |
| National prior year full eligibility population | 89.03% | 37,781,928 | 24.79% | 0.12% | 16.10% | 99.93% |
| National prior year lack full eligibility population | 87.35% | 10,621,030 | 25.48% | 0.08% | 19.18% | 99.94% |

**TABLE 7.** Shown is the model performance for the recurrent HiCC population. The threshold for a positive class is 0.92, which maximizes the F1 score. AUC is area under the ROC curve; recall is also called true positive rate or sensitivity; FPR is false positive rate; precision is also called positive predictive value; and NPV is negative predictive value.

| Population | AUC | N | Recall | FPR | Precision | NPV |
|---|---|---|---|---|---|---|
| National population | 81.23% | 28,249 | 83.28% | 36.62% | 49.98% | 89.61% |
| National prior year male population | 81.96% | 15,081 | 83.81% | 35.93% | 50.84% | 89.92% |
| National prior year female population | 80.37% | 13,168 | 82.66% | 37.40% | 49.00% | 89.25% |
| National prior year age group 0-17 population | 88.18% | 4,738 | 79.14% | 19.37% | 63.59% | 90.04% |
| National prior year age | 80.15% | 21,659 | 83.91% | 39.53% | 48.81% | 89.33% |
| group 18-64 population | | | | | | |
| National prior year age group 65+ population | 76.95% | 1,852 | 86.56% | 46.73% | 40.06% | 91.66% |
| National prior year pharmacy benefit population | 81.14% | 19,688 | 84.63% | 38.14% | 52.38% | 89.03% |
| National prior year no pharmacy benefit population | 80.65% | 8,561 | 79.08% | 33.51% | 43.37% | 90.73% |
| National prior year full eligibility population | 80.42% | 25,384 | 84.12% | 39.54% | 50.06% | 88.99% |
| National prior year lack full eligibility population | 86.33% | 2,865 | 69.37% | 15.39% | 48.37% | 93.00% |

We assessed whether the model possibly under-predicts the number of HiCCs in populations with racial minorities and found no evidence for this. In a univariate analysis correlating the average model score in each ZIP code with the ZIP's fraction of minority population we found a fairly strong positive relationship ($R^2$=47%), namely, more HiCC are predicted in areas with higher racial minority. There was no evidence of higher cost in areas of higher racial minority, and indeed the average medical cost tended to be slightly lower in these areas ($1.6 lower for every percentage point increase in minority status).

**Health economic analysis**

We estimated the financial and health impact of placing the ML algorithm in a typical care management program covering a population of 1 million individuals. In the first step of the program, identification, a subset of the population is identified as likely future high-cost claimants (Table 8). In a representative case, the program has the capacity of 1,000 HiCCs, including 500 recurrent (previously known) and 500 emergent HiCCs, respectively. We set the classification threshold separately for each population and calculated that the algorithm would attain precision of 32% and 66% for the emergent and recurrent cohorts, respectively.

**Table 8.** Shown is the ability of the machine learning algorithm to identify HiCCs for emergent and recurrent care management programs.

|  | Emergent HiCC program | | Recurrent HiCC program | |
| --- | --- | --- | --- | --- |
| Care management program capacity | Precision | True HiCCs | Precision | True HiCCs |
| 300 | 32% | 104 | 66% | 200 |
| 500 | 31% | 169 | 63% | 313 |
| 1,000 | 29% | 281 | 59% | 590 |

To compare the machine learning (ML) algorithm to a conventional rule-based system, we assumed that the program has an overall capacity of 500. The prediction threshold of the algorithm is set to 93% in order to generate 500 members with the highest risk scores. At this threshold, the precision was 39.8%, producing a population of 199 true HiCCs (Table 9). By contrast, the rule-based system identified only 10 true HiCCs, and thus the ML algorithm can impact nearly 20 times as many HiCCs. The cost of the program was $5 million in both cases, which translates to a cost per HiCC of $25,125 and $500,000 for the ML algorithm and the rule-based system, respectively. The machine learning-based system would result in a net savings of $7.3 million against a net financial cost of $4.4 million for the rule-based system.

**Table 9.** Shown is the estimated impact of HiCC identification on a typical care management program, and comparison of a conventional rule-based system with the current machine learning algorithm.

|  | Rule-based system | ML algorithm |
| --- | --- | --- |
| Members receiving case management | 500 | 500 |

| Care management program cost | $5,000,000 | $5,000,000 |
|---|---|---|
| Precision | 2% | 39.8% |
| HiCCs benefiting from interventions | 10 | 199 |
| Program cost per HiCC | $500,000 | $25,125 |
| Total savings through care management | $620,963 | $12,354,536 |
| Net savings | -$4,379,038 | $7,354,536 |

# Discussion

Our study describes an algorithm for identification of high-cost claimants at the level of $250,000 per year using the methods of machine learning. We demonstrate that using administrative claims with census data alone makes it possible to achieve AUC-ROC scores greater than 90%, even though HiCCs represent only 0.1% of the commercially insured population in the U.S. These results compare very favorably with results published in the literature, which attain performance of 80%-85%, even for populations that are easier to identify. This opens an opportunity to make interventions and achieve significant cost savings. The performance remains essentially unchanged even in populations with limited data, such as partial-year enrollment or lack of drug benefits.

Unlike previous studies that used a single ML method, we applied a modern parallel machine learning platform that considers over 50 models and automatically tuned their hyperparameters. In our experience, the best performing non-ensemble models tended to be the eXtreme Gradient Boosting algorithm [45] and its derivative, the LightGBM [47]. We also found that ensembling (or blending) of models tended to increase performance by 0.2% of AUC-ROC (results not shown here), but these models were not adopted because these gains were outweighed by the increase in computational cost and because they created certain practical barriers to model deployment.

While the literature on predictive models traditionally focuses on AUC-ROC and AUC-PR, in this application a more important measure of performance was the precision (also known as PPV) at the highest-risk 1,000 members. This is simply due to logistics and the extreme rareness of very high-cost claimants. Because the program has a limited enrollment capacity, only the highest-risk HiCCs are referred to this program, and the financial and health outcomes are influenced by the precision in this elite cohort.

The model was assessed by considering the model's precision and recall, combined with the projected effectiveness and cost of interventions. We found that in a typical care management program, the ML algorithm would create significant health and economic benefits. When compared to a typical rule-based system, the algorithm identifies approximately 20 times as many high-cost, high-needs individuals, and thus has a nearly 20 times lower cost per case. The care management program results in considerable net savings ($7.3 million) versus a net cost in the rule-based system ($4.3 million).

Future work could attempt to improve the model further, at the very least by incorporating new predictive information. Our experience indicates that claims data are an incredibly rich source of information, and we believe that there remain opportunities to improve model performance through the synthesis of new predictor variables. Another possibility for improvement might be the inclusion of external sources of information from specialty sources, such as credit reporting databases and electronic health records.

## Limitations and appropriate use of the model

A general limitation of all predictive methods in practice is that they do not give a prescriptive solution, and usually a separate prescriptive methodology is needed for planning interventions for HiCCs, which is outside the scope of this work. Common interventions include finding better and cost-efficient healthcare settings, closely monitoring members to control chronic conditions, and others. Many but not all members at risk for high costs would be willing or able to receive interventions [8].

Multiple efforts have been made to make the trained model free of errors, including quality assurance and other good software development life cycle practices. Prediction algorithms need to be regularly updated due to rapid changes in the healthcare system, including new treatments or treatment pathways and changes in costs. We plan to address these evolving changes by implementing regular updates to the training data and the model after deployment. Because our training data contains only the U.S. commercially insured population and their dependents, the model's performance is expected to degrade in populations aged 65 or older, or in populations without commercial insurance. We evaluated the performance of the model in a variety of populations and found that the degradation is small; however, the model should be used with more caution.

The algorithm presented here is primarily designed to predict the risk of high cost, rather than measures of health status or health needs.  Health is known to be systematically different from cost, and populations with barriers to healthcare, such as lower socio-economic classes or certain racial minorities, have systematically lower healthcare expenditure [55,56].  Furthermore, the algorithm uses demographic data such as age, gender, and ZIP code-linked variables in order to maximize its predictive performance.  Therefore, the most appropriate use of this model is either in a strictly financial setting, or in a holistic care management decision system.  Any use of the predictions should be performed by a healthcare professional equipped with rich contextual data because this context would allow the healthcare professional to account for contextual information not available to the algorithm, account for gaps in the algorithm's performance, and ensure equitable outcomes.

## Conclusion

The predictive model described demonstrates the potential for the next generation of predictive algorithms for the healthcare space. High-cost claimants exceeding $250,000 in annual cost account for nearly 10% of overall costs but are very rare, representing just 1.6 in every 1,000 members. By using hundreds of variables, rich claims data, and modern machine learning, it was possible to train a machine learning model that attains an AUC-PR of 91% and a precision of more than 30% in the top 1,000 members. With the high predictive performance of this model, cost-effective interventions could be implemented.

## Acknowledgments:

We thank our colleagues Michael Rogero, Munir Islam, Joulan Wu, and Carolyn Jevit, for implementing this algorithm and DataRobot for its support. Richard Paul provided oversight of the solution architecture and release management. Alan Schwartz, Ilya Safro, Ehsan Sadrfaridpour, Justin Sybrandt, and Brian Hartman provided helpful comments that greatly improved this study.

# Supporting Information

**Table S1.** The top 20 input variables of the final model ranked by variable importance. Variable importance was calculated based on the weighted number of tree splits and has been normalized so that the most important variable (AGE) has a relative importance of 1. A total of 255 variables were used in the final model. The allowed amount, sometimes simply referred to as cost, is the cost of care after the settlement between payers and providers.

| Variable | Description | Variable Importance |
|---|---|---|
| AGE | Biological age | 1.000 |
| ALLWD_AMT_FOURTH_3MO_RISING_WV | The allowed amount weighted by a linear wavelet function rising from 0 to 1 over the fourth quarter of the current year | 0.695 |
| OPTIMAL_LIFE_EXPECTANCY | Life expectancy for a healthy person of same gender/age based on actuarial tables | 0.653 |
| PREDICTED_12MO_ALLWD_AMT | The output of a submodel estimating the allowed amount in the prediction year | 0.595 |
| YLL_CURRENT_YR | Years of life lost in the current year | 0.542 |
| ALLWD_AMT_SECOND_6MO_RISING_WV | Similar to above, measuring rising costs in the second half of the current year | 0.526 |
| INPATIENT_DAYS_12MO | Number of inpatient days in the current year | 0.500 |
| ANNUAL_ALLWD_AMT_CURRENT_YEAR | Summed total allowed amounts from inpatient, professional, and pharmacy claims for the 12 months in the current period, adjusted to reflect 365 days of medical coverage | 0.495 |
| TRG_DAYS_MALIGNANCY | The number of days between the most recent cancer trigger event and the last day in the reporting period | 0.489 |
| ALLWD_AMT_PRIOR_YEAR | The sum allowed amounts from inpatient, professional, and pharmacy claims for the 12 months prior to the current period | 0.489 |
| TRG_TOTAL_ALLWD_AMT | The sum of costs associated with all trigger conditions during the reporting period | 0.479 |
| ALLWD_AMT_FALLING_WV | Similar to above, measuring falling costs in the current 12 months | 0.463 |
| ALLWD_AMT_FOURTH_3MO_WV | Total allowed amount in the fourth quarter of the current year | 0.458 |
| TOTAL_3_YEAR_ALLWD_AMT | Total allowed amount during the 36-month period (current reporting period and two prior years) | 0.458 |
| ANNUAL_ALLWD_AMT_PRIOR_YEAR | Total cost in the 12 months prior to the reporting period, adjusted to reflect 365 days of medical coverage | 0.442 |
| MORTALITY_RISK | Years of life lost (current year) divided by optimal life expectancy | 0.421 |
| ALLWD_AMT_RISING_WV | Similar to above, measuring rising costs in the current 12 months | 0.379 |
| GPI06_372000 | Loop diuretics, not otherwise specified, allowed amount over 12 months | 0.368 |
| ANNUAL_ALLWD_AMT_2_YEARS_PRIOR | Total cost in the 12 months prior to the reporting period, annualized to reflect 365 days of medical coverage | 0.368 |
| DAYS_SINCE_LAST_CLAIM | Days since the last available claim | 0.363 |